\documentclass[conference]{IEEEtran}
\IEEEoverridecommandlockouts

\usepackage{cite}
\usepackage{CJK}  
\usepackage[utf8]{inputenc}
\usepackage[bookmarks=false]{hyperref}

\usepackage{zhlipsum, lipsum}   % dummy text
\usepackage{graphicx}           % figures
\usepackage{booktabs}           % tables
\usepackage[ruled]{algorithm2e} % algorithms, list of algorithms
\usepackage{cite}               % bibliography
\usepackage{comment}

\usepackage{graphicx} % Required for inserting images
\usepackage{multirow}
\usepackage{adjustbox}
\usepackage{makecell}
\usepackage{amsmath} % 若要使用 \boldsymbol
\usepackage{bm}      % 若要使用 \boldsymbol
\usepackage{amsfonts}
\usepackage{amssymb}
\usepackage{pdfpages}

\def\BibTeX{{\rm B\kern-.05em{\sc i\kern-.025em b}\kern-.08em
    T\kern-.1667em\lower.7ex\hbox{E}\kern-.125emX}}

\newcommand{\phead}[1]
{
\noindent \textbf{#1}
}

\begin{document}
\begin{CJK}{UTF8}{bkai}

%\title{Kolmogorov-Arnold Network Knowledge Amalgamation from Multiple GNNs}
%\title{Graph-Free Learning for Social Networks: Knowledge Distillation from GNNs to Kolmogorov-Arnold Networks}
\title{Transferring Social Network Knowledge from Multiple GNN Teachers to Kolmogorov–Arnold Networks}

\author{\IEEEauthorblockN{Yuan-Hung Chao, Chia-Hsun Lu, and Chih-Ya Shen}
\IEEEauthorblockA{\textit{Department of Computer Science} \\
\textit{National Tsing Hua University}\\
Hsinchu, Taiwan \\
Email: yhchao@lbstr.cs.nthu.edu.tw, chlu@m109.nthu.edu.tw, chihya@cs.nthu.edu.tw}}

\maketitle

\begin{abstract}
% \lipsum[1-2]
%This study aims to integrate Kolmogorov-Arnold Networks (KANs) with existing Graph Neural Network (GNN) architectures, proposing three novel network structures—KGAT, KSGC, and KAPPNP—to enhance GNN performance on large-scale graph data. We utilize these KAN-based GNN architectures as teacher models and employ multi-teacher knowledge amalgamation techniques to distill the knowledge from multiple teacher models into graph-independent student models, such as KAN. The experimental results demonstrate that KAN-based GNNs exhibit superior performance in node classification tasks, and multi-teacher knowledge amalgamation significantly improves student model performance, especially in few-shot learning scenarios. These findings offer new insights and directions for future research.
Graph Neural Networks (GNNs) have shown strong performance on graph-structured data, but their reliance on graph connectivity often limits scalability and efficiency. Kolmogorov-Arnold Networks (KANs), a recent architecture with learnable univariate functions, offer strong nonlinear expressiveness and efficient inference. In this work, we integrate KANs into three popular GNN architectures—GAT, SGC, and APPNP—resulting in three new models: KGAT, KSGC, and KAPPNP. We further adopt a multi-teacher knowledge amalgamation framework, where knowledge from multiple KAN-based GNNs is distilled into a graph-independent KAN student model. Experiments on benchmark datasets show that the proposed models improve node classification accuracy, and the knowledge amalgamation approach significantly boosts student model performance. %, particularly in few-shot settings. 
Our findings highlight the potential of KANs for enhancing GNN expressiveness and for enabling efficient, graph-free inference.
\end{abstract}

\begin{IEEEkeywords}
Graph Neural Network, Knowledge Distillation, Knowledge Amalgamation, Kolmogorov–Arnold Network
\end{IEEEkeywords}

\section{Introduction}
\label{c:introduction}
% \lipsum[1-3]
% \begin{table}[t]
%     \centering
%     \begin{tabular}{ c | c c }
%     \toprule
%      cell1 & cell2 & cell3 \\
%     \midrule
%      cell4 & cell5 & cell6 \\  
%      cell7 & cell8 & cell9 \\
%     \bottomrule
%     \end{tabular}
%     \caption{An example table}
%     \label{tab:my_label}
% \end{table}
With the rapid development of deep learning technologies, Graph Neural Networks (GNNs) have demonstrated exceptional performance in processing graph-structured data~\cite{zhou2020graph}. GNNs have been widely applied in social network analysis, recommendation systems, knowledge graphs, and bioinformatics. However, the reliance of GNNs on graph structure data for inference poses challenges to their scalability and efficiency when dealing with large-scale graph data. %Additionally, the training and inference processes of GNNs are computationally intensive, making real-time applications in resource-constrained environments difficult. 
To address these limitations, researchers have proposed various improvements, including Graph Convolutional Networks (GCNs)~\cite{kipf2016semi}, Graph Attention Networks (GATs)~\cite{velivckovic2017graph}, Simple Graph Convolution (SGC)~\cite{wu2019simplifying}, and APPNP~\cite{gasteiger2018predict}. These methods enhance model expressiveness and computational efficiency through different mechanisms but still fall short of completely solving the bottlenecks of GNNs in large-scale applications. On the other hand, Kolmogorov-Arnold Networks (KANs)~\cite{liu2024kan}, as a novel neural network architecture, replace traditional fixed activation functions with learnable univariate functions, demonstrating strong non-linear representation capabilities and interpretability. KANs have proven their superior performance in tasks such as mathematical and physical function fitting, achieving high accuracy with fewer training parameters. Moreover, KANs feature lower resource consumption, avoidance of catastrophic forgetting, and greater interpretability.

Recent studies have integrated KANs into GNNs~\cite{kiamari2024gkan, zhang2024graphkan, bresson2024kagnns, de2024kolmogorov}, showing some performance improvements in experiments. However, current research mainly focuses on integrating KAN with GCN, with limited exploration of other common GNN architectures. Therefore, our goal is to combine KAN with other GNN architectures (e.g., GAT~\cite{velivckovic2017graph}, SGC~\cite{wu2019simplifying}, APPNP~\cite{gasteiger2018predict}) to further validate the advantages of KAN Layers in GNNs.

Knowledge Distillation (KD)~\cite{gou2021knowledge} is a model compression and acceleration technique that trains a smaller student model to mimic a larger, more efficient teacher model. Hinton et al. first proposed KD~\cite{hinton2015distilling}, demonstrating its effectiveness in image classification tasks. KD has also been applied to GNNs to reduce computational load and enhance inference speed~\cite{yang2020distilling, jing2021amalgamating, zhang2022graphless, zhang2022multi, yang2022contrastive}. Multi-Teacher Knowledge Amalgamation (MTACL)~\cite{yang2022contrastive} is an extension of KD, which merges knowledge from multiple teacher models into a student model. This method leverages the diversity of heterogeneous teacher models, allowing the student model to learn a broader range of knowledge from different perspectives, thus improving its generalization ability and performance.

Current research mainly focuses on integrating KAN with GCN~\cite{kiamari2024gkan, zhang2024graphkan, bresson2024kagnns, de2024kolmogorov}, with limited exploration of other common GNN architectures. Our goal is to combine KAN with other GNN architectures (e.g., GAT~\cite{velivckovic2017graph}, SGC~\cite{wu2019simplifying}, APPNP~\cite{gasteiger2018predict}) to further validate the advantages of KAN Layers in GNNs. The study by MTACL~\cite{yang2022contrastive} has limitations in the choice of student models, primarily comparing different sizes of MLP student models without attempting other structures. We aim to use KAN as the student model to explore its effectiveness in knowledge distillation tasks. Additionally, one of MTACL's conclusions is that heterogeneous teacher models (those with different structures) can distill better student models. Based on this conclusion, we introduce KAN-based GNN teacher models and combine them with common GNNs as teacher pairs to verify the effect of heterogeneous teacher models in knowledge distillation. We hope to distill more performant student models through more structurally diverse teacher combinations.

We choose KAN~\cite{liu2024kan} as the student model for the following reasons: KAN has demonstrated superior performance compared to MLP in multiple studies, indicating the potential for better performance in knowledge distillation tasks. Given the inclusion of KAN-based GNNs as teacher models, we hypothesize that using KAN as the student model will result in better learning outcomes due to structural similarities. Distilling knowledge into graph-structure-independent KAN helps optimize inference time, particularly for more complex network structures like KAN-based GNNs.

The contributions of this paper are summarized as follows.

\begin{itemize}
\item We propose three new network architectures—KGAT, KSGC, and KAPPNP—that combine the powerful non-linear representation capabilities of KAN~\cite{liu2024kan} with the advantages of GAT~\cite{velivckovic2017graph}, SGC~\cite{wu2019simplifying}, and APPNP~\cite{gasteiger2018predict}, respectively, enhancing the performance of GNNs.
\item We design a training framework based on multi-teacher knowledge amalgamation, merging knowledge from multiple teacher models and transferring it to graph-structure-independent KAN student models, thereby improving inference efficiency.
\item Through experiments, we validate the superior performance of the proposed methods on multiple benchmark datasets and test various KAN variants in knowledge distillation tasks, conducting ablation studies to analyze the effects of different components.
\end{itemize}

The innovation of this study lies in combining KAN with GNNs and leveraging multi-teacher knowledge amalgamation techniques to overcome the bottlenecks of GNNs in large-scale applications, providing new ideas and directions for future research.

In addition to its technical contributions, our research offers significant practical advantages. Our model provides more precise performance compared to competitors, achieving higher accuracy across various tasks. By using Knowledge Amalgamation, we enhance efficiency while maintaining performance, which is crucial in resource-constrained environments. Moreover, our model excels in scenarios with limited training data, maintaining strong performance. This makes our research both innovative in theory and competitive in practice.

\section{Related Works}
\label{c:related works}
% \lipsum[1-3]
% \begin{figure}
%     \centering
%     \includegraphics[width=.6\linewidth]{example-image-a}
%     \caption{An example figure.}
%     \label{fig:example-a}
% \end{figure}
% \subsection{Section Title}
% \lipsum[4-5]
% \subsection{Subsection Title}
% \lipsum[6]
\subsection{Kolmogorov-Arnold Networks (KANs)}
Kolmogorov-Arnold Networks (KANs)~\cite{liu2024kan} are a novel neural network architecture designed based on the Kolmogorov-Arnold representation theorem~\cite{kolmogorov1956representation, kolmogorov1957representation, braun2009constructive}. This theorem states that any multivariable function can be represented as a combination of several univariate functions, providing a mathematical foundation for KANs. The unique feature of KANs is the replacement of traditional linear weights with learnable spline-based univariate functions, which serve as activation functions.

KANs overcome the limitations of traditional multilayer perceptrons (MLPs) in terms of efficiency and interpretability. They reduce information loss and demonstrate high performance with low computational resource consumption and good model interpretability. Additionally, KANs perform well in continuous learning tasks, effectively avoiding catastrophic forgetting.

%Since their introduction, KANs have been applied in various fields such as computer vision~\cite{azam2024suitability}, time series forecasting~\cite{vaca2024kolmogorov}, and more. These applications highlight KANs' broad applicability in both time series and image data. Overall, KANs provide a new direction and a powerful tool for the development of neural networks through their innovative design.

\subsection{Graph Neural Networks (GNNs)}
Graph Neural Networks (GNNs)~\cite{zhou2020graph} are a specialized neural network model designed for processing graph data. Unlike traditional Neural Networks, GNNs aggregate and propagate node features through the graph structure, effectively capturing the dependencies between nodes. This type of network model has demonstrated its advantages in various fields such as social network analysis~\cite{li2023survey}, bioinformatics~\cite{zhang2021graph}, and recommendation systems~\cite{zhou2020graph}. The following are some representative GNN models and their applications:
%\phead{Graph Convolutional Networks (GCNs).}
GCNs~\cite{kipf2016semi} capture local information in graph structures by performing convolution operations on node features using adjacency matrices. %The GCN model proposed by Kipf and Welling is widely applied and has achieved excellent results on various graph datasets. GCNs effectively aggregate information from neighboring nodes, making them particularly useful for node classification and graph classification tasks.
%\phead{Graph Attention Networks (GATs).}
GATs~\cite{velivckovic2017graph} introduce an attention mechanism that adaptively assigns weights to different neighboring nodes to aggregate node features, enhancing the model's expressiveness. %The GAT model proposed by Velickovic et al. performs well in node classification and link prediction tasks. By incorporating the attention mechanism, GATs can aggregate features based on the importance of neighboring nodes, improving model flexibility and accuracy.
%\phead{Simple Graph Convolution (SGC).}
SGC~\cite{wu2019simplifying} simplifies the GCN model by removing nonlinear activation functions and combining weight matrices, thus improving computational efficiency while maintaining model performance. %SGC's design philosophy is to simplify the network structure to reduce computational overhead, achieving excellent results on most graph datasets. This makes SGC very practical in scenarios requiring efficient computation.
%\phead{Approximate Personalized Propagation of Neural Predictions (APPNP).}
APPNP~\cite{gasteiger2018predict} combines personalized PageRank with graph neural networks for efficient semi-supervised learning. %APPNP first transforms node features using a multilayer perceptron (MLP) and then performs multiple layers of propagation using personalized PageRank to achieve effective information aggregation and transmission. This method is particularly suitable for applications that need to consider both node features and graph structures, such as node classification and graph classification.

Recent studies have integrated Kolmogorov-Arnold Networks (KANs) with GNNs~\cite{kiamari2024gkan, zhang2024graphkan, bresson2024kagnns, de2024kolmogorov}, typically replacing learnable weights and fixed activation functions in traditional GCNs with KAN layers. This integration has shown significant performance improvements and broad application potential. For instance, GraphKAN~\cite{zhang2024graphkan} has demonstrated outstanding performance in few-shot classification tasks, suggesting the potential of KAN-based GNNs in few-shot learning.

\subsection{Knowledge Amalgamation Techniques}
Knowledge Amalgamation (KA) techniques aim to teach a student model by combining knowledge from multiple teacher models from different domains, leveraging the diversity of teacher models to enhance the student's learning effectiveness~\cite{shen2019amalgamating, luo2019knowledge,jing2021amalgamating}. KA techniques have been widely applied in various domains, including GNNs. Zhang et al.~\cite{zhang2022multi} proposed the Multi-Scale Knowledge Distillation (MSKD) method, applying KA to the GNN domain. They found that the accuracy of GNN models with different layers depends on node degrees. By amalgamating the topological semantics and attention mechanisms of multiple teacher models, a smaller GNN student model can further improve its performance. Moreover, Yang et al.~\cite{yang2022contrastive} developed the Multiple Teachers Amalgamation with Contrastive Learning (MTACL) method, advancing KA techniques. MTACL combines knowledge from multiple teacher models to teach a graph-independent MLP student model and introduces a contrastive learning component to enhance the student's feature representations. MTACL uses an attention mechanism to dynamically assign weights based on teacher model performance, adaptively amalgamating knowledge from different teacher models. However, MTACL has limitations in the choice of student models, mainly comparing MLP student models of different sizes without attempting other structures. Therefore, MTACL lacks verification of the effectiveness of different structured student models. Our research will further explore the use of KAN as a student model. MTACL concluded that heterogeneous teacher models (i.e., structurally different teacher models) can produce better student models. Based on this conclusion, we will introduce KAN-based GNN teacher models and combine them with common GNNs as teacher combinations to further verify the effectiveness of heterogeneous teacher models in knowledge distillation. We aim to produce more performant student models through teacher combinations with greater structural diversity. 

\section{Preliminary }
\label{c:problem definition}
% \lipsum[1-3]
% \begin{algorithm}[t]
% \SetAlgoLined
% \KwResult{Write here the result }
%  initialization\;
%  \While{While condition}{
%   instructions\;
%   \eIf{condition}{
%    instructions1\;
%    instructions2\;
%    }{
%    instructions3\;
%   }
%  }
%  \caption{How to write algorithms}
% \end{algorithm}
In this section, we introduce the concepts and architecture of Kolmogorov-Arnold Networks (KANs)~\cite{liu2024kan} and introduce existing KAN-based graph neural network (GNN) architectures.
\subsection{Kolmogorov-Arnold Networks (KANs) }
Kolmogorov-Arnold Networks (KANs)~\cite{liu2024kan} are based on the Kolmogorov-Arnold representation theorem~\cite{kolmogorov1956representation, kolmogorov1957representation, braun2009constructive}. A KAN Layer consists of multiple learnable univariate functions, parameterized using spline functions. The feature update process can be expressed as: $h^{(l+1)}_j=\sum _{i=1}^{d_l}\phi^{(l)}_{j,i}\,\,h^{(l)}_i$, 
%\begin{equation}
%    h^{(l+1)}_j=\sum _{i=1}^{d_l}\phi^{(l)}_{j,i}\,\,h^{(l)}_i
%\end{equation}
where $\phi^{(l)}_{j,i}$  denotes the spline function applied between the $i$-th output feature of layer $l$ and the $j$-th input feature of layer $(l+1)$. The entire operation of a KAN Layer can further be represented in the form of matrix multiplication: 

\begin{align*}
H^{(l+1)} 
&=\begin{bmatrix}
\phi_{1,1}^{(l)}(\cdot) & \phi_{1,2}^{(l)}(\cdot) & \cdots & \phi_{1,d_l}^{(l)}(\cdot) \\
\phi_{2,1}^{(l)}(\cdot) & \phi_{2,2}^{(l)}(\cdot) & \cdots & \phi_{2,d_l}^{(l)}(\cdot) \\
\vdots & \vdots & \ddots & \vdots \\
\phi_{d_{l+1},1}^{(l)}(\cdot) & \phi_{d_{l+1},2}^{(l)}(\cdot) & \cdots & \phi_{d_{l+1},d_l}^{(l)}(\cdot)
\end{bmatrix}H^{(l)} \\
%&=\Phi(H^{(l)})\\
%&= \text{KAN}(H^{(l)})\\
\end{align*}
The entire KAN Layer is essentially a $d_l \times d_{l+1}$ function matrix, where $d_l$ represents the feature dimension of layer $l$, and $d_{l+1}$ represents the feature dimension of layer $(l+1)$. In summary, to emphasize that the neural network layer is a KAN Layer, we use $\text{KAN}(H)$ to represent the entire KAN Layer. %This design allows KANs to exhibit superior performance in representing and learning multivariate continuous functions, especially in terms of reducing information loss and computational resource consumption~\cite{liu2024kan}.

\subsection{Existing KAN-based GNN Architectures }
Existing research has explored the combination of Kolmogorov-Arnold Networks (KANs)~\cite{liu2024kan} with Graph Neural Networks (GNNs)~\cite{zhou2020graph} to enhance feature extraction and representation capabilities for graph-structured data. Among these, Graph Kolmogorov-Arnold Networks (GKAN)~\cite{kiamari2024gkan} is a representative architecture. The primary innovation of GKAN lies in applying KANs in the feature extraction process within GNNs, specifically in the following two combination methods:

\phead{Architecture 1: Applying KAN Layer After Aggregation.}In this architecture, traditional graph convolution operations are first performed to aggregate information from neighboring nodes. Then, during the feature extraction stage, the KAN Layer replaces the traditional Multilayer Perceptron (MLP). This significantly enhances the model's nonlinear expressive power. The specific formula is as follows:
\begin{equation}
\mathbf{H}^{(l+1)} = \text{KAN}(\hat{A}\mathbf{H}^{(l)})
\end{equation}

where $\hat{A}$ is the normalized adjacency matrix, and $\Phi$ is a vector function composed of several univariate spline functions $\phi_i$.

\phead{Architecture 2: Applying KAN Layer Before Aggregation.}In this architecture, each node's features are first transformed nonlinearly using the KAN Layer, followed by the graph convolution operation to aggregate information from neighboring nodes. This design further improves feature representation capabilities. The specific formula is as follows:

\begin{equation}
\mathbf{H}^{(l+1)} = \hat{A}\text{KAN}(\mathbf{H}^{(l)})
\end{equation}

This method organically combines nonlinear feature extraction with neighborhood information aggregation, enhancing the overall performance of the model.

These KAN-based GNN architectures demonstrate superior feature extraction and representation capabilities, particularly showing significant advantages in few-shot classification tasks~\cite{zhang2024graphkan}. These research outcomes provide inspiration for our design, which will be further explored in the methodology section, where we propose KAN-based GNN architectures.

\section{Methodology}

%\subsection{KAN-based Graph Neural Networks}
In this section, we introduce three novel graph neural network architectures based on Kolmogorov-Arnold Networks (KANs): KGAT, KSGC, and KAPPNP. These architectures aim to improve the performance of graph neural networks in terms of feature representation and computational efficiency.

\subsection{KGAT Architecture}
\phead{Original GAT Architecture.}Graph Attention Networks (GATs)~\cite{velivckovic2017graph} aggregates node features by adaptively assigning different weights to neighboring nodes. The basic operations of GAT can be divided into two main steps:

\begin{enumerate}
\item \textbf{Neighborhood Information Aggregation}: In this step, each node aggregates information from its neighboring nodes. The aggregation weights are dynamically calculated by the attention mechanism, as follows:
   
   \begin{equation}
   \alpha_{ij} = \frac{\exp(\mathrm{LeakyReLU}(a^T [Wh_i || Wh_j]))}{\sum_{k \in \mathcal{N}_i} \exp(\mathrm{LeakyReLU}(a^T [Wh_i || Wh_j]))}
   \end{equation}
   
   where $h_i$ and $h_j$ are the feature vectors of nodes $i$ and $j$, $W$ is the weight matrix for feature transformation, $a$ is the weight vector for calculating attention, and $\mathcal{N}_i$ is the set of neighboring nodes of node $i$.
   
\item \textbf{Feature Update}: After aggregating neighborhood information, each node updates its feature representation as follows: $h_i' = \sigma \left( \sum_{j \in \mathcal{N}_i} \alpha_{ij} W h_j \right)$, 
%\begin{equation}
%h_i' = \sigma \left( \sum_{j \in \mathcal{N}_i} \alpha_{ij} W h_j \right)
%\end{equation}
where $\sigma$ is the activation function, typically chosen as LeakyReLU.
\end{enumerate}

\phead{Proposed KGAT Architecture.}Building on GAT, we introduce Kolmogorov-Arnold Networks (KANs)~\cite{liu2024kan} to further enhance the model's expressive power and computational efficiency. We designed two different KGAT architectures:

\begin{itemize}
\item \textbf{Architecture 1: Replacing Attention Vectors and Nonlinear Activation Functions with KAN Layer}

In this architecture, we replace the attention vectors $a^T$ and the nonlinear activation function LeakyReLU in GAT with KAN Layer to enhance the model's nonlinear expressive power. The specific formula is as follows:

\begin{equation}
\alpha_{ij} = \frac{\exp({a}_{\mathrm{KAN}}[Wh_i || Wh_j])}{\sum_{k \in \mathcal{N}_i} \exp({a}_{\mathrm{KAN}}[Wh_i || Wh_j])}
\end{equation}

\item \textbf{Architecture 2: Replacing the Linear Transformation Matrix with KAN Layer}

In this architecture, we replace the linear transformation matrix $W$ in GAT with KAN Layer while retaining the original LeakyReLU and attention vectors $a^T$. The specific formula is as follows:

\begin{equation}
\small
\alpha_{ij} = \frac{\exp(\mathrm{LeakyReLU}(a^T[\text{KAN}(h_i) || \text{KAN}(h_j)]))}{\sum_{k \in \mathcal{N}_i} \exp(\mathrm{LeakyReLU}(a^T[\text{KAN}(h_i) || \text{KAN}(h_j)]))}
\end{equation}

This architecture retains part of the original structure of GAT but enhances the model's expressive power by introducing the KAN Layer to replace the linear transformation matrix.

\end{itemize}

\subsection{KSGC Architecture}
\phead{Original SGC Architecture.}Simple Graph Convolution (SGC)~\cite{wu2019simplifying} is a method designed to simplify the structure of Graph Convolutional Networks (GCNs)~\cite{kipf2016semi}. The simplified operations include the following three steps:

\begin{enumerate}
\item \textbf{Removal of Nonlinear Activation Functions}: SGC removes the nonlinear activation functions between each layer, making the model linear and simplifying the network structure.

\item \textbf{Combining Multiple Convolution Operations}: SGC combines multiple convolution operations into a single operation, as shown in the following formula: $H^{(l+1)} = \hat{A}^k H^{(0)}$, 
%\begin{equation}
%H^{(l+1)} = \hat{A}^k H^{(0)}
%\end{equation}
where $\hat{A}$ is the regularized adjacency matrix. This operation achieves the effect of multiple convolutions through a single matrix multiplication.

\item \textbf{Weight Matrix Operation}: In SGC, the weight matrix $W$ is simplified by removing the intermediate nonlinear activation functions and combining the weight matrices between layers. The final feature representation can be expressed as:

\begin{equation}
Z = \hat{A}^k X W
\end{equation}

where $X$ is the initial input feature matrix, $W$ is the combined weight matrix, and $Z$ is the final output feature matrix.

\end{enumerate}

This structure design of SGC allows it to significantly improve computational efficiency while ensuring model performance.

\phead{Proposed KSGC Architecture.}Building on SGC, we introduce Kolmogorov-Arnold Networks (KANs)~\cite{liu2024kan} to further enhance the model's nonlinear expressive power and computational efficiency. Specifically, our KSGC architecture is as follows:

\begin{equation}
Z = \text{KAN}(\hat{A}^k X)
\end{equation}

In this architecture, we first perform the graph convolution operation $\hat{A}^k H^{(0)}$, and then input the convolution results into the KAN Layer for nonlinear transformation. The core idea of SGC is to improve computational efficiency by simplifying the network structure and removing nonlinear activation functions. After introducing the KAN Layer, KSGC, while adding a layer of nonlinear transformation at the outermost layer, still retains the design of combining multiple convolution operations. This design increases the model's nonlinear expressive power while maintaining SGC's efficient computational characteristics. Therefore, KSGC still aligns with SGC's original intention to improve model performance while maintaining computational efficiency.

\subsection{KAPPNP Architecture}

\phead{Original APPNP Architecture.}Approximate Personalized Propagation of Neural Predictions (APPNP)~\cite{gasteiger2018predict} is a graph neural network method based on personalized PageRank, designed to improve the performance of semi-supervised classification through an improved propagation scheme. The basic operations of APPNP include the following steps:

\begin{enumerate}
\item \textbf{Prediction Generation}:
The initial feature matrix \(X\) is transformed by a Multilayer Perceptron (MLP) to generate the initial prediction matrix \(H^{(0)}\):

\begin{equation}
H^{(0)} = \text{MLP}(X)
\end{equation}

\item \textbf{Personalized PageRank Propagation}:
Personalized PageRank (PPR) is used for multi-layer propagation to achieve effective aggregation and transmission of information. The specific formula is as follows:

\begin{equation}
H^{(k+1)} = (1-\alpha) \hat{A} H^{(k)} + \alpha H^{(0)}, \quad k=1, \ldots, K
\end{equation}

where $\hat{A}$ is the normalized adjacency matrix, $\alpha$ is the residual connection weight, and $K$ is the number of iterations.
\end{enumerate}

\phead{Proposed KAPPNP Architecture.}Building on APPNP, we introduce Kolmogorov-Arnold Networks (KANs)~\cite{liu2024kan} to further enhance the model's nonlinear expressive power and computational efficiency. First, The initial features \(X\) are transformed by the KAN Layer into \(H^{(0)}\):

\begin{equation}
H^{(0)} = \text{KAN}(X)
\end{equation}

Next, multi-layer graph convolution operations are performed through Personalized PageRank propagation. This step maintains the original design.

This design retains the efficient information aggregation and transmission characteristics of APPNP while introducing the nonlinear expressive power of KAN, thereby enhancing the model's performance.

\subsection{Multi-Teacher Knowledge Amalgamation}

Our multi-teacher knowledge amalgamation method is based on the MTACL framework~\cite{yang2022contrastive} with modifications to enhance the performance of the student model during the knowledge distillation process. First, we compute the output $\mathbf{Z}^t$ for each pre-trained teacher model. Next, we calculate the attention weight $\alpha_t$ for each teacher model based on the similarity between the teacher and student output embeddings. The formula is as follows:
\begin{equation}
\alpha_t = \frac{\exp(\mathrm{sim}(\mathbf{Z}^t, \mathbf{Z}^s))}{\sum_{k=1}^T \exp(\mathrm{sim}(\mathbf{Z}^k, \mathbf{Z}^s))}
\end{equation}
where the similarity calculation is given by: $\mathrm{sim}(\mathbf{Z}^t, \mathbf{Z}^s) = \frac{1}{n^2} \sum_{i=1}^n \sum_{j=1}^n A^t_{ij}$, 
%\begin{equation}
%\mathrm{sim}(\mathbf{Z}^t, \mathbf{Z}^s) = \frac{1}{n^2} \sum_{i=1}^n \sum_{j=1}^n A^t_{ij}
%\end{equation}
and $A^t$ is defined as: $A^t = (\mathbf{Z}^s)^T \cdot \mathbf{Z}^t$. 
%\begin{equation}
%A^t = (\mathbf{Z}^s)^T \cdot \mathbf{Z}^t
%\end{equation}
Then, we fuse the output embeddings of the teachers into a super teacher by computing the weighted average of the teacher models' outputs $\mathbf{z}_i$ as the soft target:

\begin{equation}
\mathbf{z}_i = \text{softmax}\left(\sum_{t=1}^T \alpha_t \mathbf{Z}^t_{[i,:]}\right)
\end{equation}

Next, we calculate the knowledge amalgamation loss, which is the KL divergence between the weighted average output of the teacher models and the student model's output. We aim for the student's output to approximate the teacher's:

\begin{equation}
\mathcal{L}_{MTA} = \frac{T^2}{|V|} \sum_{v \in V} \mathcal{L}_{KL} \left(\frac{\hat{y}_v}{T}, \frac{\mathbf{z}_v}{T}\right)
\end{equation}

The original MTACL framework uses multi-teacher model knowledge amalgamation and contrastive learning to enhance the student model's performance. However, in our experiments, contrastive learning did not significantly improve performance, so we simplified the framework by removing contrastive learning and instead used the commonly employed Hard target loss in Knowledge Distillation~\cite{hinton2015distilling}, which is the KL divergence between the student model and the true labels:

\begin{equation}
\mathcal{L}_{hard} = \sum_{v \in V} \mathcal{L}_{KL}(y_v, \mathbf{z}^s_v)
\end{equation}

Finally, the overall loss function combines the knowledge amalgamation loss $\mathcal{L}_{MTA}$ and the Hard target loss:

\begin{equation}
\mathcal{L}_{total} = \lambda \mathcal{L}_{MTA} + (1 - \lambda) \mathcal{L}_{hard}
\end{equation}

This design leverages the advantages of multi-teacher model knowledge amalgamation and the simplified hard target loss, effectively enhancing the performance and training efficiency of the student model.

\section{Experiments Results}
\label{c:experiments}
In this section, we compare the functionality-preserving and robustness of our proposed approach with other state-of-the-art baselines on two benchmark datasets and two model architectures.

\subsection{Experimental Settings}

\phead{Datasets.}We conduct our experiments on three commonly used graph datasets, which are widely applied in graph neural network research. The basic information of these datasets is summarized in the Table \ref{table:dataset_characteristics}.

\begin{table}[h]
\centering
\caption{Dataset Characteristics}
\label{table:dataset_characteristics}
\begin{tabular}{@{}lcccc@{}}
\toprule
Dataset   & Nodes  & Edges   & Features & Classes \\ \midrule
Cora      & 2,485  & 5,069   & 1,433    & 7       \\
Citeseer  & 2,110  & 3,668   & 3,703    & 6       \\
AMZ-P     & 7,487  & 119,043 & 745      & 8       \\ \bottomrule
\end{tabular}
\end{table}

Both Cora and Citeseer~\cite{sen2008collective} are classic citation network datasets where nodes represent papers and edges denote citation relationships. Node features are based on a bag-of-words model, with Cora having 1,433 features and Citeseer 3,703 features, divided into seven and six classes respectively. Amazon-Photo (AMZ-P)~\cite{shchur2018pitfalls} is a subset of the Amazon product dataset, where nodes represent products and edges denote co-purchase relationships. Node features are based on product descriptions, with 745 features divided into eight classes.

We focus on the largest connected component in the graph. We randomly assigned 20 labeled nodes from each category to the training set, another 30 nodes from each category for validation, and all remaining nodes to the test set.

\phead{Model Setup.}In our experiments, we used ten teacher models, including four commonly used GNN models (GCN~\cite{kipf2016semi}, GAT~\cite{velivckovic2017graph}, SGC~\cite{wu2019simplifying}, APPNP~\cite{gasteiger2018predict}) and their KAN variants (GKCN\_arch1, GKCN\_arch2, KGAT\_arch1, KGAT\_arch2, KSGC, KAPPNP). Each model has 2 layers with 32 hidden units. For GAT and KGAT variants, we set the number of attention heads to 4. For SGC and KSGC, we used K=2, and for APPNP and KAPPNP, we used K=10 and alpha=0.1. KAN variants incorporated a KAN Layer with grid\_size=32 and spline\_order=3.

For student models, both MLP and KAN were configured with 2 layers and 32 hidden units. The KAN model used a KAN Layer with grid\_size=32 and spline\_order=3, ensuring a fair comparison in the knowledge distillation process.

All experiments, including the training of teacher models and the knowledge distillation training, are performed by averaging the results of 10 runs to ensure the stability and reproducibility of the results.

\phead{Evaluation Metrics.}We used accuracy, which measures the proportion of correctly classified nodes in the test dataset, and inference time, the average time required for the model to make predictions on the test dataset, measured in milliseconds, to evaluate model performance.

\subsection{Performance of KAN Based GNNs}

We first evaluated all the GNN teacher models on the node classification task. The performance of each model on the Cora, Citeseer, and Amazon Photo datasets is summarized in the Table \ref{tab:gnn_performance}.

\begin{table}[h]
\caption{Performance of different GNN architectures on Cora, CiteSeer, and Amazon Photo datasets.}
\label{tab:gnn_performance}
\centering
\begin{tabular}{lccc}
\hline
\textbf{GNNs} & \textbf{Cora} & \textbf{CiteSeer} & \textbf{AMZ-P} \\ \hline
KGCN\_arch1   & 0.818         & 0.723             & 0.909                 \\
KGCN\_arch2   & 0.817         & 0.719             & 0.910                 \\
KGAT\_arch1   & 0.819         & 0.726             & 0.915                 \\
KGAT\_arch2   & 0.814         & 0.722             & 0.908                 \\
KSGC          & 0.802         & 0.724             & 0.902                 \\
KAPPNP        & 0.814         & 0.726             & 0.913                 \\
GCN           & 0.808         & 0.706             & 0.905                 \\
GAT           & 0.812         & 0.703             & 0.903                 \\
SGC           & 0.814         & 0.718             & 0.900                 \\
APPNP         & 0.811         & 0.724             & 0.906                 \\ \hline
\end{tabular}
\end{table}

As shown in the results, KAN Based GNNs generally outperform traditional GNNs. These KAN-based models surpass their baseline models across all three datasets. Among them, KGAT\_arch1 achieved the highest accuracy on all datasets.

These results validate the hypothesis that introducing the KAN Layer can further improve the performance of various GNN models. The powerful nonlinear representation capabilities of the KAN Layer enable these models to better capture the complex structures in graph data, thereby enhancing the accuracy of the classification tasks. This demonstrates that integrating KAN with models such as GCN, GAT, SGC, and APPNP can lead to significant performance improvements.

A better teacher model is expected to produce a better student model. Therefore, we introduce KAN Based GNNs as teacher models to provide an opportunity to obtain better student models. In this experiment with teacher models, since KGCN\_arch1 performed better than KGCN\_arch2, and KGAT\_arch1 performed better than KGAT\_arch2, we will use KGCN\_arch1 to represent KGCN and KGAT\_arch1 to represent KGAT in the subsequent knowledge distillation experiments.

\subsection{Multi-Teacher Knowledge Amalgamation}

In this section, we combine teacher models in pairs to perform Knowledge Amalgamation, distilling their knowledge into a KAN student model. We aim to observe which combinations can improve the student's performance. The Table \ref{tab:multi} shows the performance of each combination on the Cora, Citeseer, and Amazon Photo datasets.

\begin{table}[h]
\caption{Performance of multi-teacher knowledge amalgamation on Cora, CiteSeer, and Amazon Photo datasets.}
\label{tab:multi}
\centering
\begin{tabular}{lccc}
\hline
\textbf{Teacher1, Teacher2} & \textbf{Cora} & \textbf{CiteSeer} & \textbf{AMZ-P} \\
\hline
APPNP, SGC & 0.817 & 0.721 & 0.904 \\
GAT, APPNP & 0.814 & 0.734 & 0.911 \\
GAT, GCN & 0.810 & 0.725 & 0.918 \\
GAT, SGC & 0.816 & 0.729 & \textbf{0.918} \\
GCN, APPNP & 0.812 & 0.732 & 0.914 \\
GCN, SGC & 0.812 & 0.717 & 0.915 \\
KAPPNP, APPNP & 0.815 & 0.728 & 0.908 \\
KAPPNP, GAT & 0.819 & 0.732 & 0.915 \\
KAPPNP, GCN & 0.814 & \textbf{0.732} & 0.913 \\
KAPPNP, SGC & 0.817 & 0.728 & 0.906 \\
KGAT, APPNP & \textbf{0.819} & 0.736 & 0.919 \\
KGAT, GAT & 0.816 & \textbf{0.736} & 0.918 \\
KGAT, GCN & 0.814 & 0.734 & \textbf{0.921} \\
KGAT, KAPPNP & 0.818 & 0.734 & 0.916 \\
KGAT, SGC & 0.818 & 0.732 & 0.918 \\
KGCN, APPNP & 0.815 & 0.723 & 0.911 \\
KGCN, GAT & 0.816 & 0.724 & 0.912 \\
KGCN, GCN & 0.810 & 0.727 & 0.917 \\
KGCN, KAPPNP & \textbf{0.819} & 0.730 & 0.911 \\
KGCN, KGAT & 0.818 & 0.730 & 0.916 \\
KGCN, SGC & \textbf{0.819} & 0.724 & 0.909 \\
KSGC, APPNP & \textbf{0.820} & 0.733 & 0.904 \\
KSGC, GAT & \textbf{0.820} & \textbf{0.740} & 0.914 \\
KSGC, GCN & \textbf{0.827} & 0.724 & 0.912 \\
KSGC, KAPPNP & 0.822 & \textbf{0.740} & 0.906 \\
KSGC, KGAT & 0.821 & 0.731 & 0.916 \\
KSGC, KGCN & 0.818 & 0.729 & 0.907 \\
KSGC, SGC & 0.814 & 0.735 & 0.900 \\
\hline
\end{tabular}
\end{table}

We observe that when the student model learns from multiple teachers, the overall accuracy improves compared to single-teacher knowledge distillation. The best combinations for each dataset are as follows: for Cora, KSGC combined with GCN achieves an accuracy of 0.827; for Citeseer, KSGC combined with GAT achieves an accuracy of 0.741; and for Amazon Photo, KGAT combined with GCN achieves an accuracy of 0.921.

A notable phenomenon is that the best combinations in all datasets involve one KAN Based GNN paired with a traditional GNN. Additionally, the KAN Based GNN's base model is different from the paired traditional GNN. For example, in the Cora dataset, the best combination is KSGC with GCN, rather than with its base model SGC.

These results verify our previous hypothesis that combinations of teacher models with greater structural differences can lead to better student model performance. This allows the student model to learn knowledge from different perspectives, enhancing its overall capability. Therefore, we strategically select diverse teacher model combinations in the knowledge amalgamation process to maximize the performance improvement of the student model.

\section{Conclusion}
\label{c:conclusion}
%This study introduces three novel KAN-based GNN architectures—KGAT, KSGC, and KAPPNP—and designs a training framework based on multi-teacher knowledge amalgamation. Our experimental results show that these KAN-based GNNs outperform traditional GNNs in node classification tasks. Additionally, our knowledge amalgamation technique effectively enhances student model performance by combining the knowledge from structurally diverse teacher models, validating the advantage of heterogeneous teacher model combinations in knowledge distillation.

%We also find that distilling knowledge into graph-independent KAN student models not only improves inference efficiency but also maintains high classification accuracy. This indicates that KAN has potential advantages as a student model in knowledge distillation tasks.

%In summary, this research provides new directions for the design of GNNs and demonstrates the potential of multi-teacher knowledge amalgamation techniques in improving student model performance. These results offer significant insights for efficient inference in large-scale graph data applications and pave the way for further research.
We proposed three KAN-based GNN architectures—KGAT, KSGC, and KAPPNP—and a multi-teacher knowledge amalgamation framework for distilling their knowledge into a graph-independent KAN student model. Experiments on benchmark datasets show that the proposed models improve node classification performance and that heterogeneous teacher combinations lead to better student models. The results also demonstrate that KAN offers an efficient and effective alternative for graph-free inference. These findings suggest promising directions for scalable and flexible graph learning.

\bibliographystyle{IEEEtran}
\bibliography{references}

\end{CJK}
\end{document}